\documentclass[conference]{IEEEtran}
\usepackage{amsmath}
\usepackage{graphicx}
\usepackage{hyperref}
\usepackage{algorithm}
\usepackage{algpseudocode}
\usepackage{multirow}

\newtheorem{theorem}{Theorem}

\newtheorem{proposition}[theorem]{Proposition}

\def\BibTeX{{\rm B\kern-.05em{\sc i\kern-.025em b}\kern-.08em
    T\kern-.1667em\lower.7ex\hbox{E}\kern-.125emX}}
\begin{document}

%%%%%%%%%%%%%%%%%%%%%%%%%%%%%%%%%%%%%%%%%%%%%%%%%%%%%%%%%%%%%%%%%%%%%%%%

% \begin{frontmatter}

%%% Use this command to specify your submission number.
%%% In doubleblind mode, it will be printed on the first page.

% \paperid{123} 

%%% Use this command to specify the title of your paper.

\title{Vanishing Variance Problem in Fully Decentralized \\ Neural-Network Systems}

%%% Use this combinations of commands to specify all authors of your 
%%% paper. Use \fnms{} and \snm{} to indicate everyone's first names 
%%% and surname. This will help the publisher with indexing the 
%%% proceedings. Please use a reasonable approximation in case your 
%%% name does not neatly split into "first names" and "surname".
%%% Specifying your ORCID digital identifier is optional. 
%%% Use the \thanks{} command to indicate one or more corresponding 
%%% authors and their email address(es). If so desired, you can specify
%%% author contributions using the \footnote{} command.

% \author[A]{\fnms{Yongding}~\snm{Tian}\orcid{0000-0002-4515-4009}\thanks{Corresponding Author. Email: Y.Tian-3@tudelft.nl}}
% \author[A]{\fnms{Zaid}~\snm{Al-Ars}\orcid{0000-0001-7670-8572}% \thanks{Corresponding Author. Email: Z.Al-Ars@tudelft.nl}
% }
% \author[A]{\fnms{Maksim}~\snm{Kitsak}\orcid{0000-0002-6862-6021}}
% \author[A,B]{\fnms{Peter}~\snm{Hofstee}\orcid{0000-0001-9649-7338}} 

% \address[A]{Delft University of Technology, Delft, NL}
% \address[B]{IBM Infrastructure, Austin, TX, US}

\author{
\IEEEauthorblockN{Yongding Tian\IEEEauthorrefmark{1}, Zaid Al-Ars\IEEEauthorrefmark{1}, Maksim Kitsak\IEEEauthorrefmark{1}, Peter Hofstee\IEEEauthorrefmark{2}.}
\IEEEauthorblockN{Email: \{Y.Tian-3, Z.Al-Ars\}@tudelft.nl},

\IEEEauthorblockA{\IEEEauthorrefmark{1}\textit{Delft University of Technology, Delft, NL}}
\IEEEauthorblockA{\IEEEauthorrefmark{2}\textit{IBM Infrastructure, Austin, TX, US}}

}

%%% Use this environment to include an abstract of your paper.

\maketitle

\begin{abstract}

% gpt pass,
Federated learning and gossip learning are emerging methodologies designed to mitigate data privacy concerns by retaining training data on client devices and exclusively sharing locally-trained machine learning (ML) models with others. The primary distinction between the two lies in their approach to model aggregation: federated learning employs a centralized parameter server, whereas gossip learning adopts a fully decentralized mechanism, enabling direct model exchanges among nodes. This decentralized nature often positions gossip learning as less efficient compared to federated learning. Both methodologies involve a critical step: computing a representation of received ML models and integrating this representation into the existing model. Conventionally, this representation is derived by averaging the received models, exemplified by the FedAVG algorithm. Our findings suggest that this averaging approach inherently introduces a potential delay in model convergence. We identify the underlying cause and refer to it as the "vanishing variance" problem, where averaging across uncorrelated ML models undermines the optimal variance established by the Xavier weight initialization. Unlike federated learning where the central server ensures model correlation, and unlike traditional gossip learning which circumvents this problem through model partitioning and sampling, our research introduces a variance-corrected model averaging algorithm. This novel algorithm preserves the optimal variance needed during model averaging, irrespective of network topology or non-IID data distributions. Our extensive simulation results demonstrate that our approach enables gossip learning to achieve convergence efficiency comparable to that of federated learning.

\end{abstract}

% \end{frontmatter}

%%%%%%%%%%%%%%%%%%%%%%%%%%%%%%%%%%%%%%%%%%%%%%%%%%%%%%%%%%%%%%%%%%%%%%%%

\section{Introduction}

% gpt pass,
Google introduced \textit{federated learning} as a solution to privacy concerns associated with the aggregation of user data for training machine learning (ML) models \cite{McMahan2016CommunicationEfficientLO, Enthoven2021}. In a typical federated learning scenario, a central parameter server disseminates a global model to client devices. These devices then update the global model based on their local data and return the updated model to the server for aggregation. This decentralized approach mitigates privacy problems by ensuring that client devices only share ML models, rather than raw data, with external parties, thus keeping sensitive data localized. %Given that federated learning is designed for model training on decentralized and distributed data, federated learning is also recognized as a method for \textit{decentralized learning} and \textit{distributed learning}.

% gpt pass,
At the same time, \textit{gossip learning} emerged as an alternative framework for decentralized ML, eliminating the need for a central server, while opting for direct peer-to-peer exchanges among client devices \cite{https://doi.org/10.1002/cpe.2858}. This mode of operation positions gossip learning as a more decentralized approach compared to federated learning. Nonetheless, the absence of a central server in gossip learning has been identified as a factor that may impair convergence performance when contrasted with federated learning, as documented in existing research \cite{HEGEDUS2021109, 10.1007/978-3-030-22496-7_5}.

% gpt pass,
A common challenge encountered in both federated learning and gossip learning is the minimization of communication costs. To address this , existing research has proposed various solutions, including knowledge distillation \cite{wu_communication-efficient_2022}, model compression techniques \cite{Sattler2019Robust, Yang2022Online} for federated learning, and model partitioning/sampling strategies \cite{HEGEDUS2021109, 10.1007/978-3-030-22496-7_5} for gossip learning. Intriguingly, our findings indicate a delay of convergence in gossip learning when model compression methods are not employed. This outcome is counter-intuitive, as model compression methods are typically lossy and associated with diminished learning efficiency.

% gpt pass,
Our investigation of this delay of convergence reveals that Glorot et al.\ first described such a phenomenon as a "plateau" in single node training \cite{pmlr-v9-glorot10a}. To circumvent this plateau delay, Glorot et al.\ developed the "Xavier optimal variance" for weight initialization, proposing a methodology to configure model weights accordingly. This weight initialization strategy effectively mitigates the plateau delay in single-node settings, and has become the standard for non-bias weight initialization. In the context of gossip learning, the absence of a central server leads to each node independently initializing their model weights. However, these independently initialized, uncorrelated weight distributions can result in reduced variance following model averaging, thereby compromising back-propagation efficiency and causing the plateau delay.

% gpt pass,
To address the plateau delay, we propose an additional step of re-scaling weight values to adhere to the Xavier optimal variance after model averaging. This re-scaling succeeds in eliminating the plateau delay. At the same time, this adjustment proves to be inconsequential for correlated models, such as those aggregated by a central server in federated learning, ensuring compatibility with existing model averaging frameworks. Our simulation results suggest that this modification enables gossip learning to achieve comparable convergence performance as federated learning. We call this combined approach, including the original model averaging technique, as variance-corrected model averaging.

% gpt pass,
The contributions of this paper are summarized as follows:
\begin{itemize}
    \item We identify a pervasive phenomenon, termed "plateau delay", in gossip learning systems. Existing studies primarily focus on mitigating this delay rather than eliminating it. 
    \item We identify the cause of the plateau delay as the vanishing variance problem that occurs when averaging uncorrelated neural network models.
    \item We propose a variance-corrected averaging algorithm to address the vanishing variance problem.
    \item Our findings indicate that, with the implementation of the variance-corrected averaging algorithm, gossip learning can achieve comparable convergence efficiency as federated learning.
\end{itemize}

% gpt pass,
The layout of this paper is as follows: Section~\ref{sec:background} provides an overview of federated learning, gossip learning, and Xavier weights initialization. %Mentioning transfer learning here because it is another viable strategy to bypass the plateau delay. 
Section~\ref{sec:plateau_effect} discusses the gossip learning system, the plateau delay, methodologies for quantifying plateau delays, and shows how existing federated learning, gossip learning, and transfer learning can circumvent this . Section~\ref{sec:vanishing_variance} identifies the vanishing variance as the root cause of the plateau delay. Then, we present our proposed solution, called variance-corrected model averaging, in Section~\ref{sec:solution_to_vanishing_variance}. Section~\ref{sec:result} shows the performance results of implementing variance-corrected model averaging. Finally, Section~\ref{sec:conclusion} summarizes this paper.

%%%%%%%%%%%%%%%%%%%%%%%%%%%%%%%%%%%%%%%%%%%%%%%%%%%%%%%%%%%%%%%%%%%%%%%%

\section{Background}
\label{sec:background}

\subsection{Federated Learning}

% gpt pass,
Federated learning, initially proposed in \cite{McMahan2016CommunicationEfficientLO}, aims to address privacy concerns by allowing data to remain on client devices, with only model weights or gradients shared with a central server. This approach is characterized by two key entities: a parameter server and numerous client devices. The parameter server is responsible for aggregating model updates, while client devices focus on training the model with their local data before transmitting the updated model back to the server. The operational workflow of federated learning is shown in Algorithm~\ref{alg:federated_learning}.

\begin{algorithm}[htbp]
\caption{Federated Learning Procedure. Variable $w$ denotes model weights, $t$ represents the round of federated averaging.} 
\label{alg:federated_learning}
\begin{algorithmic}[5]
\Procedure{ServerExecutes}{}
    \State $w_0 \gets \text{initialize}$ \Comment{Initial model weights}
    \For{\text{$t = 0, 1, 2, \dots$}}
        \State $S_t \gets \text{availableClients}(t)$ \Comment{Clients available at time $t$}
        \For{\text{each client $k$ in $S_t$ \textbf{in parallel}}}
            \State $w^k_{t+1} \gets \text{ClientUpdate}(k, w_t)$
        \EndFor
        \State $w_{t+1} \gets \textit{Average}(\{w^k_{t+1}\}_{k \in S_t})$
    \EndFor
\EndProcedure

\State

\Function{ClientUpdate}{$k, w_t$} \Comment{Run on client $k$}
    \State $D_{train} \gets \text{collectData()}$
    \State $w_{t+1} \gets \textit{Training}(w_t, D_{train})$
    \State \textbf{return} $w_{t+1}$
\EndFunction

\end{algorithmic}
\end{algorithm}

% gpt pass,
The \textit{Training} step in this algorithm typically employs backpropagation using Stochastic Gradient Descent (SGD) to update model weights based on local data. The subsequent \textit{Average} step computes the aggregate of all updated model weights, with the simplest form being \(w_{t+1} = \frac{1}{N} \sum_{k \in S_t} w^k_{t+1}\), where \(N\) is the number of clients. The original averaging algorithm introduced by \cite{McMahan2016CommunicationEfficientLO} incorporates weights to account for the variable number of local training samples on client devices, defined as \(w_{t+1} = \sum_{k \in S_t} \frac{n_k}{m_t} w^k_{t+1}\), where \(m_t = \sum_{k \in S_t} n_k\) represents the total number of data samples across all participating clients, and \(n_k\) denotes the number of samples on client \(k\).

\subsection{Gossip Learning}

% gpt pass,
While federated learning offers a novel approach to preserving privacy by keeping data on client devices, it introduces new vulnerabilities such as data poisoning \cite{pmlr-v108-bagdasaryan20a}, \cite{9194010}, model poisoning \cite{10.5555/3489212.3489304}, privacy attacks \cite{Enthoven2021}, \cite{10062088}, and the risk of a single point of failure. In response, some researchers have proposed the removal of the central server to adopt a fully decentralized learning mechanism known as "gossip learning" \cite{HEGEDUS2021109} \cite{https://doi.org/10.1002/cpe.2858}. Gossip learning eliminates the central server, thereby mitigating the single point of failure and limiting the potential impact of poisoning attacks within a vast peer-to-peer network. Additionally, the increased network distance in gossip learning complicates privacy attacks.

% gpt pass,
In gossip learning,  clients communicate directly with their peers instead of a central server, as depicted in Algorithm~\ref{alg:gossip_learning}. The main difference from federated learning lies in the redistribution of models post-training to peers, not to a central authority. Unlike some studies that adopt a globally initialized model in gossip learning \cite{9245248}, this methodology eschews this to avoid introducing any form of centralization. Peers maintain a model buffer, aggregating received models using the \textit{Average} function and subsequently integrating this aggregated model into their own using the \textit{Update} function.

\begin{algorithm}[htbp]
\caption{The gossip learning procedure \cite{HEGEDUS2021109}. The variable $w$ represents model weights.} 
\label{alg:gossip_learning}
\begin{algorithmic}[5]
\Procedure{NodeExecutes}{}
    \State $w_0 \gets \text{initialize}$ \Comment{Performed on each node}
    \While{}
        \State $D_{train} \gets \text{collectData()}$
        \State $w \gets \textit{Training}(w, D_{train})$ 
        \State $w_s \gets \text{Sample}(w)$ \Comment{Weights sampling}
        \State $S_p \gets \text{selectPeer}()$
        \For{\text{each node $k$ in $S_p$ \textbf{in parallel}}}
            \State send $w_s$ to $k$, trigger onReceiveModel
        \EndFor
    \EndWhile
\EndProcedure

\State

\Function{onReceiveModel}{$w_s$} \Comment{Run on node $k$}
    \State $B \gets w_s$ \Comment{Store $w_s$ to buffer $B$}
    \If{$B \text{ is full}$} 
        \State $w_m \gets \textit{Average}(B)$
        \State $w \gets \textit{Update}(w, w_m)$
    \EndIf 
    
\EndFunction

\end{algorithmic}
\end{algorithm}

% gpt pass,
For the \textit{Average} step, \cite{HEGEDUS2021109} initially applied the same \textit{Average} step as federated learning and observed poor convergence performance. To improve this, they introduced sophisticated weight sampling and partitioning techniques.\footnote{Discussed in Section 3, "Gossip Learning," paragraph two.} Despite these enhancements, challenges persisted in achieving satisfactory convergence with non-independent and identically distributed (non-IID) data. The study concluded that:
% \footnote{Summarized by editor as "paper highlights," available at \url{https://www.sciencedirect.com/science/article/pii/S0743731520303890}}

\begin{enumerate}
\item Fully decentralized machine learning is a viable alternative to federated learning.
\item Compression is essential for all the algorithms to achieve competitive performance.
\item Uneven class-label distribution over the nodes favors centralization.
\end{enumerate}

% gpt pass,
Additionally, SGD-based gossip learning algorithms, like those presented in \cite{DBLP:journals/tmlr/IssaidEB22, DBLP:journals/corr/abs-2105-09080}, differ by sharing gradients with peers instead of model weights. Due to their fundamentally different approach, this paper will not cover such variations of gossip learning.

% \subsection{Transfer learning}

\subsection{Xavier Initialization}
\label{subsec:xavier_initialization}

% initial version generated by chatgpt
Xavier initialization is a method proposed by Glorot et al.\ to address the  of vanishing and exploding gradients that can occur during the training of deep neural networks \cite{pmlr-v9-glorot10a}. They state that the weights of neural networks should be initialized in such a way that the variance of the outputs is equal to the variance of the inputs. This balance helps in maintaining a steady flow of gradients through the network, which in turn makes the network easier to train. To achieve this, Xavier initialization sets the weights of a layer by drawing them from a distribution with zero mean and a specific variance that depends on the number of input and output neurons.

% initial version generated by chatgpt
Specifically, for a layer with $n_{\text{in}}$ input neurons and $n_{\text{out}}$ output neurons, the weights are initialized from a uniform distribution in the range $\left[-\frac{\sqrt{6}}{\sqrt{n_{\text{in}} + n_{\text{out}}}}, \frac{\sqrt{6}}{\sqrt{n_{\text{in}} + n_{\text{out}}}}\right]$, or a normal distribution with mean 0 and variance $\frac{2}{n_{\text{in}} + n_{\text{out}}}$ \cite{PyTorchDocs}. This approach has been shown to significantly improve the training speed and performance of deep neural networks by facilitating more efficient back-propagation of errors.

%%%%%%%%%%%%%%%%%%%%%%%%%%%%%%%%%%%%%%%%%%%%%%%%%%%%%%%%%%%%%%%%%%%%%%%%

\section{Plateau Delay in Fully Decentralized Neural-Network Systems}
\label{sec:plateau_effect}

\subsection{Experiment Setup}
\label{subsec:experiment_setup}

% gpt pass,
This paper starts with the examination of a simple gossip learning architecture, referred to as the "baseline" network. This network consists of 50 nodes, where each node is connected to 8 other peers, forming a regular graph topology with a degree of 8, denoted as $G=Regular(N=50, k=8)$. In this setup, nodes perform a training session by processing a batch of training samples every $T_{acquisition}$ ticks, a unit of virtual time scale within our system. For analytical simplicity, $T_{acquisition}$ is set to 10, making the network operation synchronous, despite gossip learning inherently operating in an asynchronous manner. Upon the completion of training, model weights are sent directly to all connected peers, thus bypassing the \textit{Sample} and \textit{selectPeer} steps in Algorithm~\ref{alg:gossip_learning}. 
We combine the \textit{Average} and \textit{Update} functions, using the following equation once the model buffer reaches capacity:

\begin{equation}
\label{equation:model_averaging_algorithm}
model_{new} = \beta \cdot model_{old} + (1-\beta) \cdot \frac{\sum_{n=1}^{N}model_n}{N}
\end{equation}

% gpt pass,
In this formula, $\beta$ is a weighted factor that dictatates the portion of the old model, $model_{old}$, used in the averaging process, thereby controlling the influence exerted by incoming models from peers, $model_n$. Here, $n$ is the index of the models within the buffer, and $N$ is the total number of peer models aggregated. For the baseline network configuration, we assign $\beta=0.5$ and $N=8$, corresponding to the network's design where each node is connected to 8 peers ($k=8$). Consequently, each node performs model averaging every 10 ticks, aligning with the training session. The interval ratio, denoted as $R = \frac{T_{averaging}}{T_{training}}$, quantifies the frequency of averaging relative to training. Thus, for our baseline network $R=1$.

% gpt pass,
To select the machine learning model used for nodes, we take both computational resources and convergence performance  into account. Therefore, we selected LeNet-5 as the model of choice \cite{LeNet}. This model adopts the hyper-parameters from Caffe's implementation of LeNet\footnote{Hyper-parameters include a base learning rate of 0.01, momentum of 0.9, weight decay of 0.0005, an inverse learning rate policy, gamma set to 0.0001, and power to 0.75. 
% Detailed configuration available at \url{https://github.com/BVLC/caffe/blob/master/examples/mnist/lenet_solver.prototxt}
}. The MNIST dataset \cite{deng2012mnist}, a collection of handwritten digit images, was utilized to train the LeNet-5 model. A batch size of 64 was set, with each node obtaining 64 randomly selected training samples from the MNIST dataset every $T_{acquisition}$ ticks for a training session. Consequently, in the baseline network configuration, each node processes 640 training samples (0.0106 epoch) every 100 ticks, totaling 32,000 training samples (0.533 epoch) network-wide.

\begin{figure}[htbp]
    \centering
    \includegraphics[width=0.99\columnwidth]{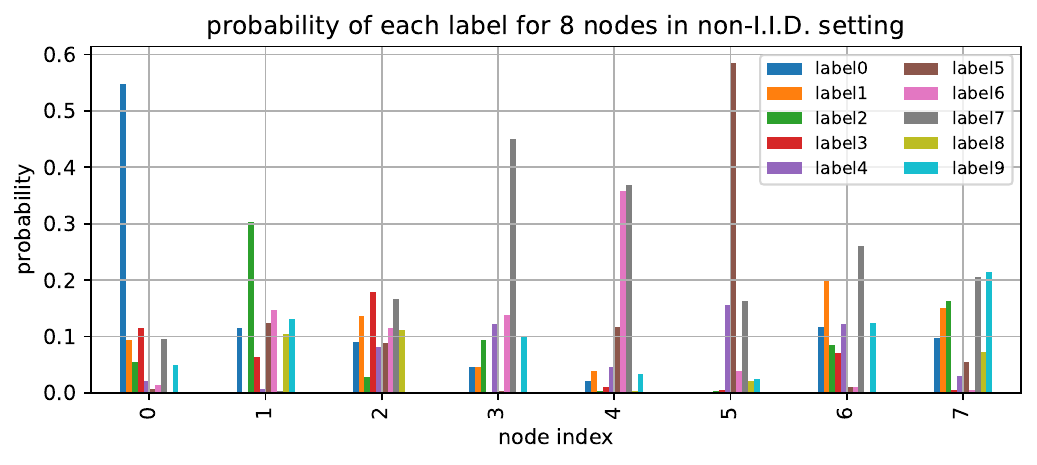}
    \caption{Label distribution under $\alpha=0.5$ for 8 nodes using the MNIST dataset, which contains 10 distinct labels. The y-axis represents the probability of each label occurring in a node's training set.}
    \label{fig:gamma_distribution}
\end{figure}

% gpt pass,
To evaluate the convergence performance in non-Independent and Identically Distributed (non-IID) datasets, we employed a symmetric Dirichlet distribution, as discussed in \cite{WU2024121390}, creating a probabilistic label distribution for each node. The Dirichlet distribution's concentration parameter, $\alpha$, reflects the dataset's IID characteristics; lower $\alpha$ values signify a stronger non-IID tendency. For the baseline network, we simulate both IID and non-IID (with $\alpha=0.5$) scenarios. Figure~\ref{fig:gamma_distribution} presents an intuitive visualization of the label distribution across 8 nodes under this non-IID setting.

% gpt pass,

% model weight difference
% gpt pass,
This paper employs two metrics to analyze the process of gossip learning. The first metric measures the accuracy of the neural network model at each node. It is expected that, owing to peer-to-peer communication, the accuracy levels across nodes will closely align, displaying only minor variations. The second metric is the model weight difference, calculated as follows:
\begin{equation}
\label{equation:model_weight_difference}
    \mathit{diff} = \frac{\sum_{n=0}^{N-1} {| model_{(n+1) \bmod N } - model_{n \bmod N} |} }{N} \nonumber
\end{equation}
where $model_n$ signifies the model at node $n$, and $N$ is the total node count. This metric computes the Manhattan distance between models layer-wise weights. The model weight difference is an array because the subtraction is performed layer-wise. Figure~\ref{fig:baseline}(b) shows an example model weight difference, which contains four curves because LeNet-5 has four layers with trainable weights.

\subsection{Plateau Delay in Gossip Learning}
\begin{figure}[htbp]
    \centering\includegraphics[width=0.99\columnwidth]{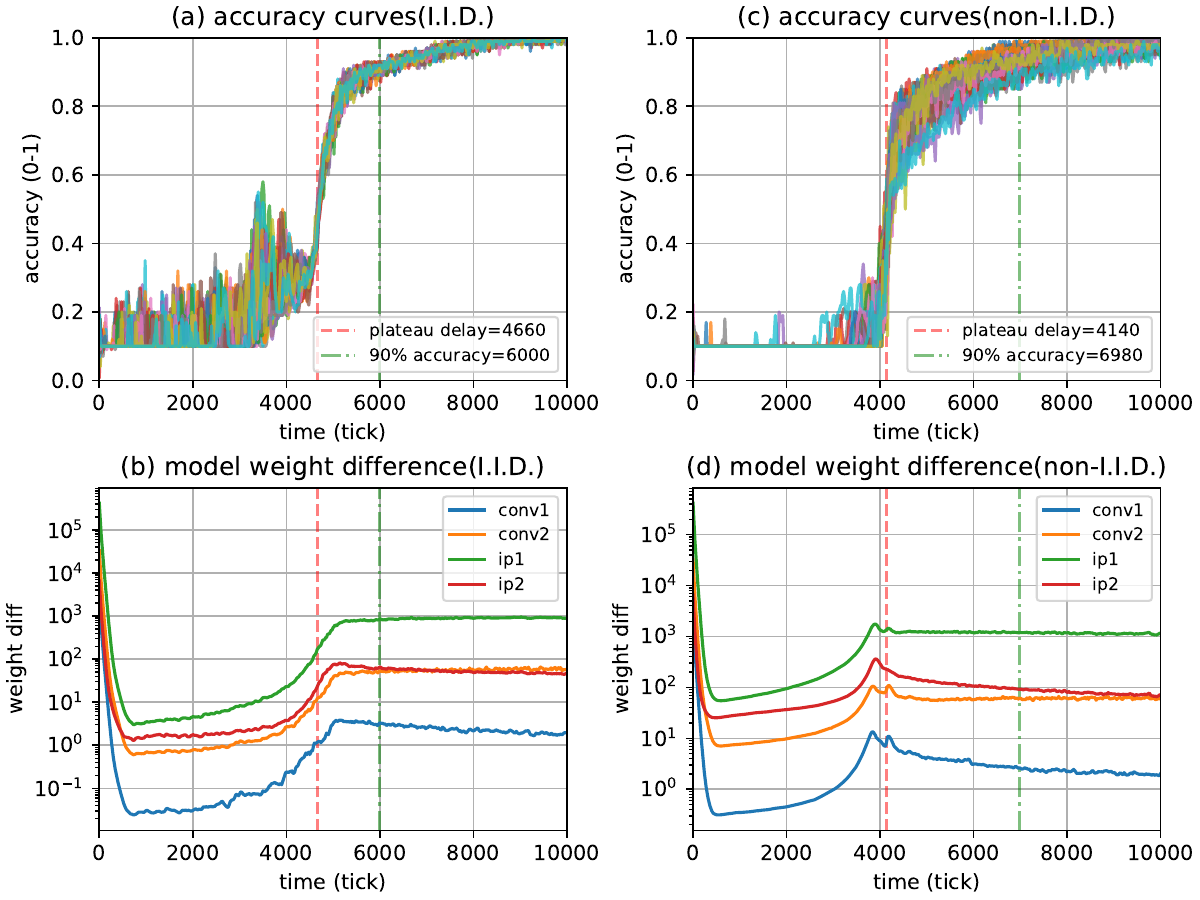}
    \caption{Accuracy curves and model weight differences for the baseline network under IID (a,b) and non-IID (c,d) settings. The "plateau delay" is calculated according to Equation~\ref{equation:definition_plateau_delay}, and "90\% accuracy" marks the point when over 90\% of nodes achieve an accuracy higher than 0.9. The "conv1", "conv2", "ip1" and "ip2" are the name of four trainable layers in LeNet.}
    \label{fig:baseline}
\end{figure}

% gpt pass,
Performing gossip learning in the baseline network across IID and non-IID settings using the DFL simulator \cite{10.1145/3600225} yielded the results shown in Figure~\ref{fig:baseline}.

% The model weight difference decreases rapidly before 1000 ticks because their models are independently initialized and gossip communications quickly align these models. 

% gpt pass,
The accuracy curves presented in Figure~\ref{fig:baseline} reveal a clear convergence delay in the model, which we refer to as the "plateau delay". While the accuracy curve of the LeNet model training on a single node typically exhibits its most rapid increase at the beginning of training, this rapid increase is delayed in our simulations. Thus, we could quantify this delay as the time difference between the start of the training and the most rapid increase, calculated by Equation~\ref{equation:definition_plateau_delay}.

\begin{equation}
\label{equation:definition_plateau_delay}
A'(t_{plateau\_delay}) \geq A'(t) \ \ \ \ \  \forall t \in (t_{first\_average}, +\infty)
\end{equation}

% gpt pass,
Here, \(A(t)\) signifies the average accuracy across all nodes at time \(t\), with \(A'(t)\) representing its derivative. The \(t_{first\_average}\) is the first model averaging time; this excludes the initial accuracy rise from isolated training. This distinction ensures the plateau delay accurately reflects the impact of gossip learning, not the early gains from training alone.

% gpt pass,
The phenomenon of plateau delay has been documented in various studies on gossip learning, such as in the observations presented in Figure 4 of \cite{9245248} and Figure 6 of \cite{10.1007/978-3-030-22496-7_5}. Notably, this delay is absent in the system described by \cite{HEGEDUS2021109}, where the major difference is the implementation of a model weights sampling step, as detailed in Algorithm~\ref{alg:gossip_learning} line 6. Furthermore, \cite{HEGEDUS2021109} emphasizes the critical role of compression in enhancing the algorithm's efficiency, leading to the hypothesis that compression may mitigate the plateau delay. This assumption challenges the conventional wisdom that model compression primarily serves to minimize bandwidth usage, rather than to speed up convergence.

% gpt pass,
Considering the absence of the plateau delay in both federated learning and weights-compressed gossip learning systems, we adapted our baseline network to mirror the architecture of these systems. This adjustment aims to evaluate the impact of such a configuration on the occurrence of the plateau delay.

\subsection{Existing Methods without Plateau Delay}

\subsubsection{Federated Learning Variant}
\label{subsub:federated_learning}
\begin{figure}[htbp]
    \centering
    \includegraphics[width=0.99\columnwidth]{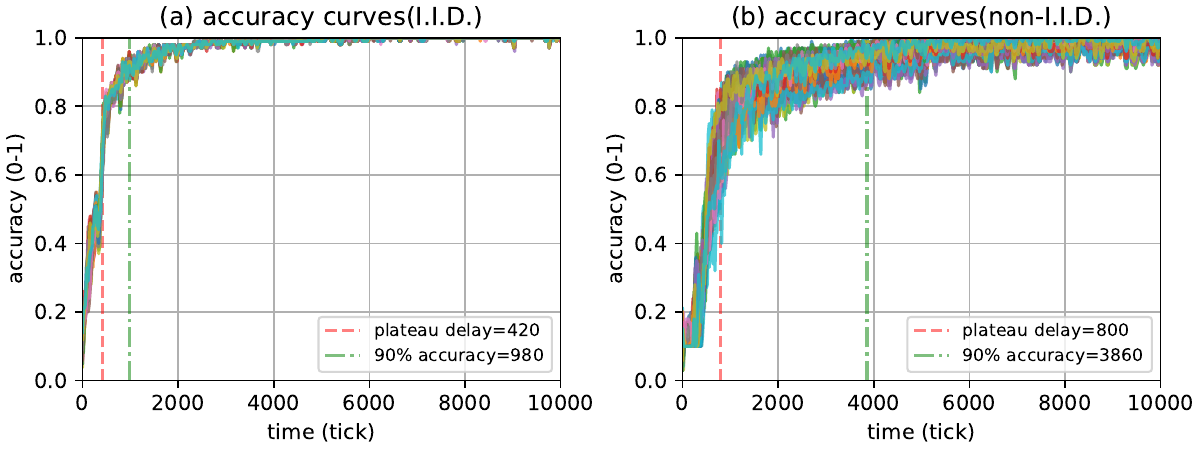}
    \caption{Accuracy curves for the baseline configuration with $Star(N=50)$ topology under IID\ (a) and non-IID\ (b) settings.}
    \label{fig:star}
\end{figure}

% gpt pass,
We adapted the network topology of our baseline from a $Regular(N=50, k=8)$ network to a star network, denoted as $Star(N=50)$. This adjustment aligns our network more closely with federated learning configurations, albeit with two notable differences: 1) our system uses the averaging algorithm defined in Equation~\ref{equation:model_averaging_algorithm} instead of FedAvg, and 2) nodes initialize their models independently. The simulation outcomes are depicted in Figure~\ref{fig:star}.

% gpt pass,
Figure~\ref{fig:star} illustrates that centralizing the decentralized system can eliminate plateau delay in the IID setting, underscoring why such delays are not observed in federated learning due to centralized aggregation. Nevertheless, a brief plateau delay persists in the non-IID setting, where reaching 90\% accuracy is significantly delayed, highlighting the challenge of training ML models with non-IID datasets \cite{Sattler2019Robust, WU2024121390}.

\subsubsection{Weights-Compressed Gossip Learning}
\label{subsubsec:compressed_gossip}
\begin{figure*}[htbp]
    \centering
    \includegraphics[width=0.8\textwidth]{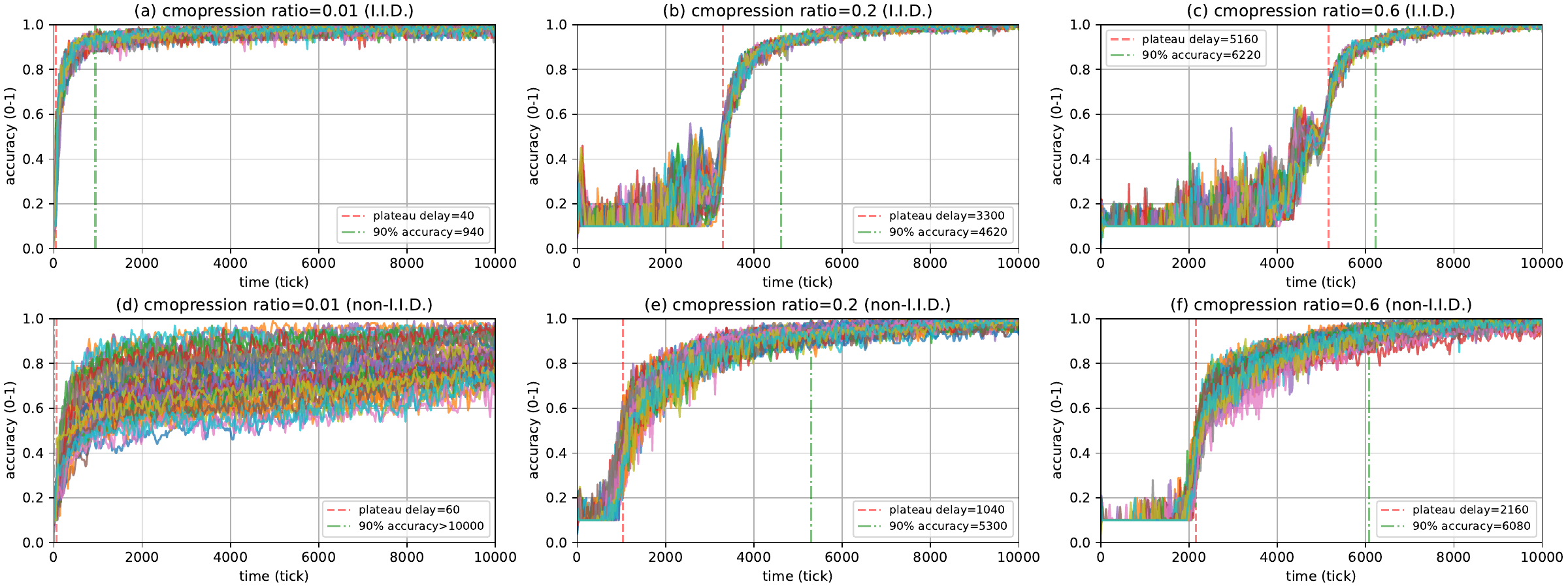}
    \caption{The accuracy curves for compression ratio=0.01, 0.2 and 0.6 in IID and non-IID settings.}
    \label{fig:compress_ratio}
\end{figure*}

% gpt pass,
To explore the impact of model weights compression, as described in \cite{HEGEDUS2021109}, we implemented a simple form of compression: sharing a partial weights with peers. We denote the extent of this sharing as the compression ratio; for instance, a compression ratio of 0.2 means only 20\% of the weights are transmitted, with selection being random. Figure~\ref{fig:compress_ratio} presents the accuracy outcomes for compression ratios of 0.01, 0.2, and 0.6 in both IID and non-IID settings.

% gpt pass,
Figure~\ref{fig:compress_ratio} reveals that model compression mitigates plateau delay and significantly enhances convergence performance in IID settings. However, under non-IID conditions (Figure~\ref{fig:compress_ratio}(d)(e)(f)), compression improves IID convergence at the expense of deteriorating performance in non-IID scenarios. The optimal IID convergence coincides with the poorest non-IID performance.

% gpt pass,
These findings align with \cite{HEGEDUS2021109}, which posits that: 1) compression is essential for all the algorithms to achieve competitive performance, and 2) uneven class-label distribution over the nodes favors centralization. Although the compression techniques in \cite{HEGEDUS2021109} outperform our simplistic random approach in convergence efficiency, the overall trends remain consistent.

\subsubsection{Transfer Learning Variant}
\label{subsubsec:transfer_learning}
% gpt pass,
Another scenario that circumvents the plateau delay involves models reaching a high level of training before averaging, similar to the transfer learning concept where pre-trained models are fine-tuned for new problems. In gossip learning, this translates to nodes only engaging with peers if their models are sufficiently trained. This typically occurs when communication is infrequent enough that a node achieves high accuracy on its dataset before engaging in communications.

% gpt pass,
To adapt our baseline network to incorporate a transfer learning-like approach, we designed a temporal hierarchical network with specific rules:

\begin{itemize}
    \item Initially, nodes have no connections; peers specified in the baseline are placed on a waiting list.
    \item A node may connect to an additional peer from the waiting list once its accuracy on its own dataset reaches 0.8.
    \item The model buffer size adjusts dynamically with the number of peers to maintain a constant training-to-averaging interval ratio ($R$).
\end{itemize}

\begin{figure}[htbp]
    \centering
    \includegraphics[width=0.8\columnwidth]{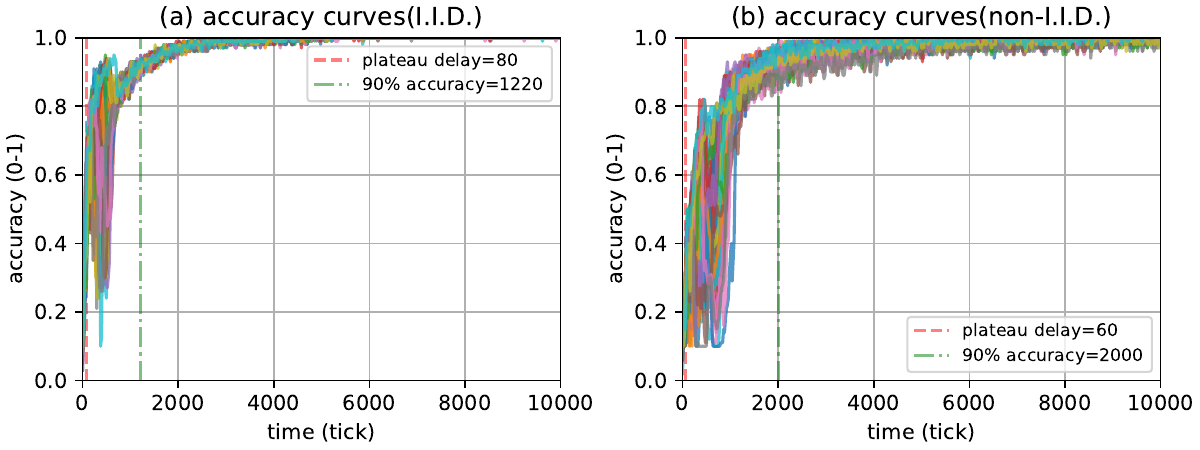}
    \caption{Accuracy curves for the temporal hierarchical network under IID (a) and non-IID (b) settings.}
    \label{fig:temporal}
\end{figure}

% gpt pass,
Ultimately, each node is expected to surpass 0.8 accuracy and establish connections with all peers on the waiting list, mirroring the baseline network's topology. Figure~\ref{fig:temporal} displays the training accuracy curves in this temporal hierarchical network.

% gpt pass,
As expected, this approach effectively mitigates the plateau delay. This strategy, including its variants such as token-based flow control \cite{HEGEDUS2021109}, significantly complicates the system and does not guarantee each node adherence to this protocol, especially in a decentralized gossip environment characterized by autonomous operation.

\section{Vanishing Variance Causes Plateau Delay}
\label{sec:vanishing_variance}

% gpt pass,
If we want to keep gossip learning fully decentralized by removing all centralized elements, such as centralized topologies and sharing the same initial model weights, then it becomes difficult to match federated learning performance without resorting to complicated methods for decentralized environment. This limitation reduces the interest researchers dedicate to gossip learning compared to federated learning.

\begin{figure}[htbp]
    \centering
    \includegraphics[width=0.6\columnwidth]{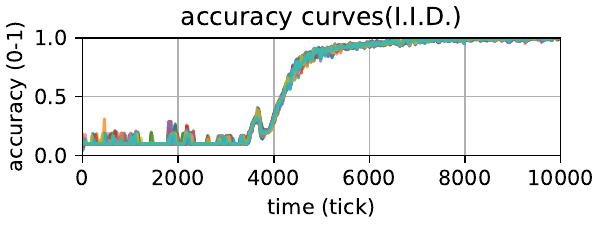}
    \caption{Accuracy curves for the star baseline network with global broadcasting mechanism.}
    \label{fig:global_average}
\end{figure}

% gpt pass,
To overcome this limitation, we investigated an unanswered critical question: why does the plateau delay occur? Through extensive experimentation across various configurations and network topologies, we were able to identify the root cause of the delay. %conclusive evidence emerged within the experiments described in Section \ref{subsub:federated_learning}. 
Repeating the experiments described in Section~\ref{subsub:federated_learning} while introducing a global broadcasting mechanism at the beginning of the model training, whereby an averaged model is calculated and uniformly applied to all nodes, led to the reemergence of the plateau delay. As elaborated in Figure~\ref{fig:global_average}, this phenomenon mirrors the delay initially noted in the baseline network as shown in Figure~\ref{fig:baseline}.

% gpt pass,
No plateau delay was noted in the system outlined in Section~\ref{subsub:federated_learning}. Therefore, the plateau delay shown in Figure~\ref{fig:global_average} comes from the global broadcasting mechanism implemented at the beginning of the simulation. Since no training precedes this global broadcasting, this indicates that the averaging process itself triggers the plateau delay. Further examination revealed that Glorot et al.\ first documented this delay in a single-node training context \cite{pmlr-v9-glorot10a}, suggesting that for effective neural network training, initial weights must be set with a specific variance. However, the \textit{Average} step in Algorithm~\ref{alg:gossip_learning} can reduce this variance in scenarios where the models are not entirely correlated. This reduction in variance for uncorrelated Xavier initialization (both normal and uniform distributions) can be quantified by the following formula:

\begin{equation}
    \label{eq:variance_decrease_in_averaging}
    \text{Var}(\overline{X}) = \frac{1}{N^2} \sum_{i=1}^{N} \sigma_i^2
\end{equation}

\begin{figure}[htbp]
    \centering
    \includegraphics[width=0.8\columnwidth]{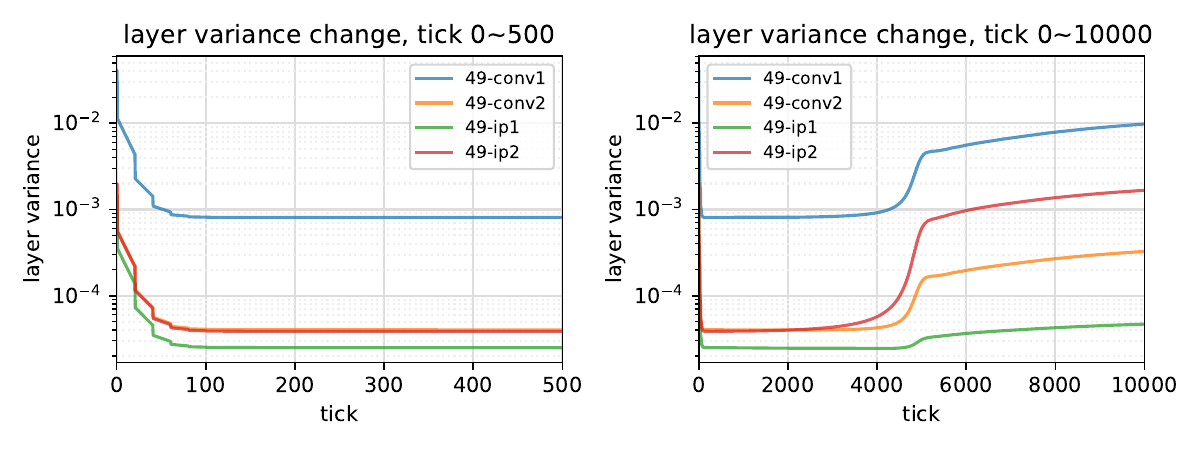}
    \caption{Variance curves for the baseline network.}
    \label{fig:baseline_variance}
\end{figure}

% gpt pass,
This formula indicates that the variance, \(\text{Var}(\overline{X})\), of the averaged distribution \(\overline{X}\) is the sum of the variances of individual distributions \(X_i\), scaled down by \(N^2\), where \(N\) is the number of distributions and \(\sigma_i\) is the standard deviation of distribution \(X_i\). It is important to note that post-averaging, models tend to exhibit increased correlation, thereby decreasing the rate of variance reduction compared to the \(N^2\) factor. To illustrate this effect, we analyzed the variance of layer weights within the system depicted in Figure~\ref{fig:baseline} and plotted the variance curve for node 49. The results, shown in Figure~\ref{fig:baseline_variance}, confirm a significant reduction in layer variance—nearly 50-fold within the initial 100 ticks.

% gpt pass,
We define this reduction in variance as the "vanishing variance problem," which also explains how the previously discussed variations of the baseline network effectively counter this problem:
\begin{itemize}
    \item Section~\ref{subsub:federated_learning}: the star network architecture, where the central node disseminates its model across the network, ensures the predominance of the central model. This dominance staves off a rapid decrease in variance. However, in scenarios involving global broadcasting (where averaging induces variance reduction) the plateau delay reemerges. Gossip learning implementations that initialize with the same model across nodes similarly mitigate plateau delays by correlating initial models.
    \item Section~\ref{subsubsec:compressed_gossip}: model compression slows the rate of variance decrease, thereby diminishing the plateau delay.
    \item Section~\ref{subsubsec:transfer_learning}: postponing the averaging process until after models have achieved high accuracy helps to preserve variance during training, and preventing the diminishing variance effect of averaging. The strategy of incrementally connecting peers effectively reduces the value of \(N\) in Equation~\ref{eq:variance_decrease_in_averaging}, further mitigating variance reduction.
\end{itemize}

% gpt pass,
The discussion in this section is summarized by the following proposition.

\begin{proposition}
\label{prop:model_averaging_reduce_variance}
Model averaging among uncorrelated neural network models leads to reduced model weights variance, which can impede subsequent model training efforts.
\end{proposition}

\section{Variance-Corrected Model Averaging}
\label{sec:solution_to_vanishing_variance}

% gpt pass,
To counteract the reduction of weight variance in Proposition~\ref{prop:model_averaging_reduce_variance}, we propose an innovative step, as shown in Algorithm \ref{alg:variance_maintained_model_averaging}, to maintain weight variance. This step involves rescaling the weights of the averaged model to match the average variance of all contributing models before averaging. This correction is crucial to apply on a layer-by-layer basis, considering that the Xavier initialization sets a different variance for each layer.

\begin{algorithm}[htbp]
\caption{Variance-Corrected Model Averaging Algorithm.} 
\label{alg:variance_maintained_model_averaging}
\begin{algorithmic}[5]

\Function{VarCorrected-ModelAveraging}{$B$} \Comment{Model buffer $B$}
    \State $\textit{Model}_{avg} \gets Average(B)$ \Comment{Compute average model}
    \State $N \gets \text{len}(B)$
    \For{$\text{Layer $l_{avg}$ in $\textit{Model}_{avg}$}$}
        \For{$\textit{Model}_i$ in $B$}
            \State $\sigma_{li}^2 \gets \textit{Var}(l_{i})$  \Comment{Variance of layer $l$ in model $i$}
        \EndFor
        \State $\sigma_{l}^2 = \frac{1}{N} \sum_{i=1}^{N} \sigma_{li}^2$
        \State $\sigma_{lavg}^2 \gets \textit{Var}(l_{avg})$
        
        \For{Each weight $v$ in $l_{avg}$}
            \State $v \gets (v - \overline{l_{avg}})\frac{\sigma_{l}}{\sigma_{lavg}} + \overline{l_{avg}}$ \Comment{Rescale to correct variance}
        \EndFor
    \EndFor
    \State \Return $\textit{Model}_{avg}$
\EndFunction

\end{algorithmic}
\end{algorithm}

% gpt pass,
% As detailed in Algorithm~\ref{alg:variance_maintained_model_averaging}, 
This algorithm ensures that the variance of the averaged model ($\textit{Model}_{avg}$) equals the mean variance of the input models ($\textit{Model}_i$). This strategy mitigates the variance reduction  identified in Proposition~\ref{prop:model_averaging_reduce_variance} for uncorrelated machine learning models. For correlated models, such as the models recevied by a federated learning server, the impact of the variance rescaling step is minimal, as the ratio $\frac{\sigma_{l}}{\sigma_{lavg}}$ approaches 1, indicating minimal variance deviation.

\section{Results}
\label{sec:result}

\subsection{%Performance 
Comparison with FL and GL} %of Federated Learning, Gossip Learning and Variance-Corrected Gossip Learning Systems}

% gpt pass,
This section shows the convergence efficiency of federated learning and gossip learning frameworks using Algorithm~\ref{alg:variance_maintained_model_averaging} for the \textit{Average} step. To compare both systems fairly, we matched the number of training nodes: the federated system with 50 clients and one central server, and the gossip system with 50 nodes as per our baseline setup described in Section~\ref{subsec:experiment_setup}. In addition, the averaging weighting factor ($\beta$) in the model averaging algorithm (Equation~\ref{equation:model_averaging_algorithm}) is set to 0 for gossip learning, to reflect the model replacement strategy in federated learning. However, a non-zero $\beta$ is also considered to account for security concerns within gossip learning environments, where node trustworthiness cannot be guaranteed. Simulations were thus also conducted with $\beta=0.5$, corresponding to the baseline network's settings.

\begin{figure}[htbp]
    \includegraphics[width=0.99\columnwidth]{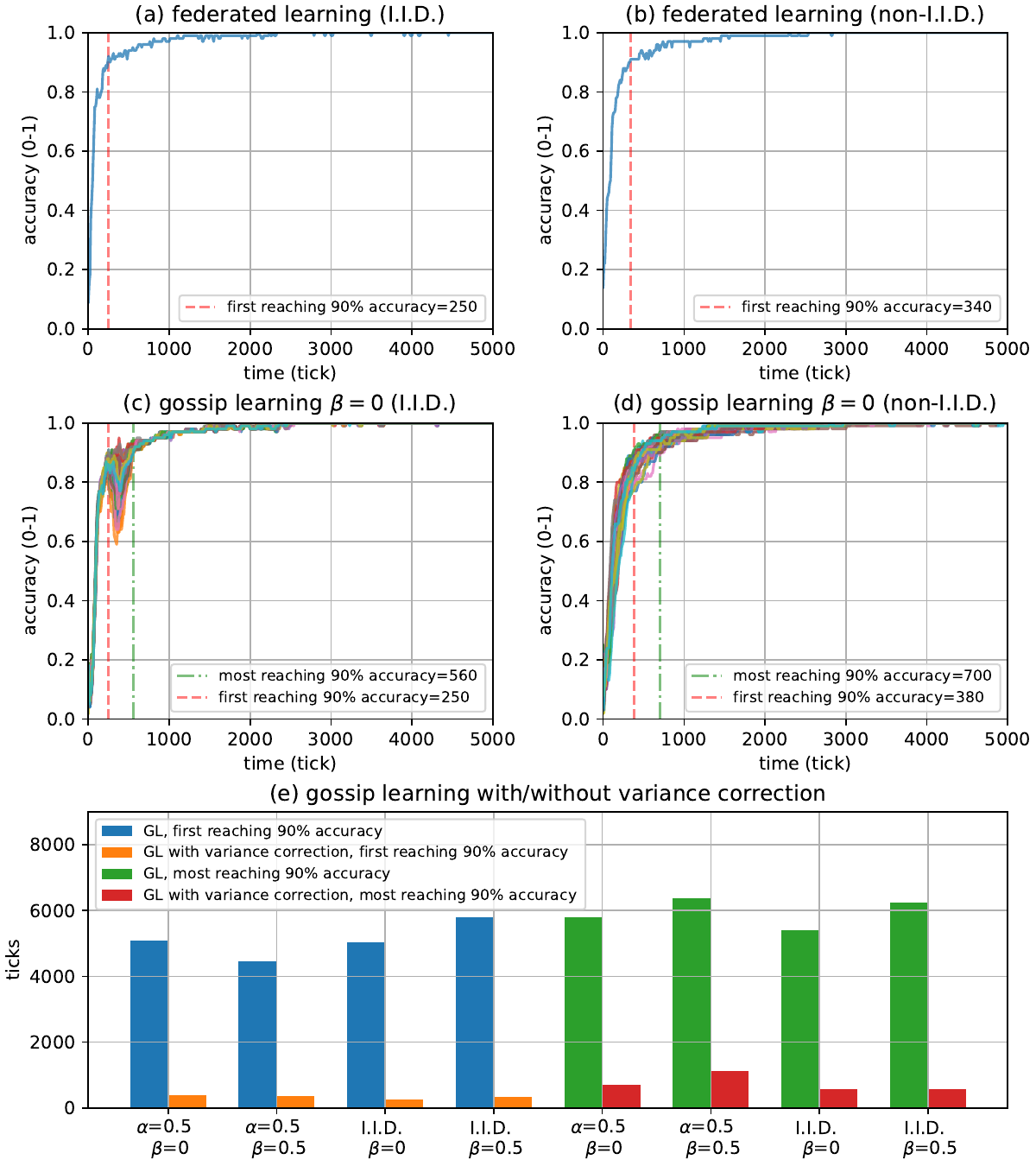}
    \caption{Convergence comparison between federated learning (a)(b), gossip learning with variance correction ($\beta=0$) (c)(d). Subfigure (e) compares the convergence performance with and without applying variance correction in gossip learning.}
    \label{fig:final_fl_gl_comparision}
\end{figure}

% gpt pass,
The comparison of convergence rates between federated learning and gossip learning with variance correction is depicted in Figure~\ref{fig:final_fl_gl_comparision}. For federated learning, we only display the accuracy of the central server. The legend item "First reaching 90\% accuracy" denotes the time point for the first node to achieve 0.9 accuracy, while "most reaching 90\% accuracy" signifies the time point where over 90\% of nodes reach 90\% accuracy. The parameter $\alpha=0.5$ in (b)(d)(e) represents the non-IID level, as shown in Figure~\ref{fig:gamma_distribution}. %When comparing scenarios in Figure~\ref{fig:final_fl_gl_comparision}(a)(c) (IID) and (b)(d) (non-IID), we observe that both learning methods enable the fastest node to reach 0.9 accuracy almost simultaneously.
The figure shows that for both IID (Figure~\ref{fig:final_fl_gl_comparision}(a)(c)) and non-IID (Figure~\ref{fig:final_fl_gl_comparision}(b)(d)) that our variance correction approach improves learning efficiency significantly, allowing gossip learning to catch up with federated learning convergence efficiency.

% ???!!!???!!!???!!

% This is a negative formulation. Please compare first the convergence of gossip learning with and without Algorithm 3. then you show a bar chart with only convergence times for all simulation settings FL, GL,  IID non-IID, B=0, B=0.5. $\beta=0,0.5$ ????!!!!!???!!

% gpt pass,
Figure~\ref{fig:final_fl_gl_comparision}(e) compares the time taken to reach "first reaching 90\%" and "most reaching 90\%" for gossip learning without variance correction (referred to as "GL" in the legend) and gossip learning with variance correction (referred to as "GL with variance correction"). We could see that applying variance correction could achieve much faster convergence time, nearly 10x faster.

% gpt pass,

\subsection{Comparison with GL Variants} %of Variance Correction with Weight Compression and Transfer Learning Variant}

We compare the performance of variance correction with the combination of weight compression (Section \ref{subsubsec:compressed_gossip}) and the transfer learning (TL) variant (Section \ref{subsubsec:transfer_learning}). To explore broader scenarios, we modified the following parameters based on the baseline network:

\begin{itemize}
    \item The network was expanded to encompass 1000 nodes ($N=1000$) to evaluate its scalability.
    \item We increased $R$ from 1 to 4 to investigate the impact of less frequent model averaging on convergence for non-IID data.
    \item The parameter $k$ was adjusted from 8 to 32, to investigate the influence of more peers. 
    \item The training interval, $T_{acquisition}$, is set to follow a uniform distribution ranging from 1 to 19. This adjustment approximates the asynchronous nature of gossip learning.
\end{itemize}

We simulated all combinations of the above modifications and present the results in Table~\ref{tab:final_results_after_modification}. The left four columns indicate the values of the modified parameters, and the right four columns present the time taken for "most reaching 90\%" for the two methods for IID and non-IID data. Our findings suggest that variance correction achieves smaller convergence times than its competitor in all cases. Additionally, variance correction exhibits better scalability, 
with up to 6x ($T_{acqu}$$=$$10, R$$=$$4, k$$=$$8$) faster convergence compared to its competitor in $N$$=$$1000$ cases.
% with convergence times ranging approximately from 560 to 1130 ticks, whereas its competitor already reaches 2600 to 4470 ticks.
This is because traditional model averaging results in further decrease in variance for larger networks, as indirectly indicated by the increasing $N$ in Equation \ref{eq:variance_decrease_in_averaging}.

Further simulations of gossip learning with only weights compression or only with TL show certain cases where "most reaching 90\%" time exceeds 10000 ticks. Due to this poor performance, we do not present these results in Table~\ref{tab:final_results_after_modification}. The cause of such poor performance of weight compression is the increasing network size and insufficient communication caused by compression in non-IID cases. For TL, reduced communication frequency and non-IID data cause nodes to reach high accuracy on their own data making them establish peer connections too rapidly, reducing both accuracy and variance after averaging. This prevents the system from escaping the plateau delay.

\begin{table}[htbp]
\centering
\caption{The time of "most reaching 90\%" under various modifications for the two gossip learning systems.}
\label{tab:final_results_after_modification}
\resizebox{\columnwidth}{!}{
\begin{tabular}{c|c|c|c|cc|cc}
\hline
\multirow{2}{*}{$T_{acqu}$} &
  \multirow{2}{*}{R} &
  \multirow{2}{*}{N} &
  \multirow{2}{*}{k} &
  \multicolumn{2}{l|}{\begin{tabular}[c]{@{}l@{}}GL with TL and \\ weights compression\end{tabular}} &
  \multicolumn{2}{l}{\begin{tabular}[c]{@{}l@{}}GL with variance \\ correction\end{tabular}} \\ \cline{5-8} 
     &   &      &    & \multicolumn{1}{l|}{IID} & $\alpha=0.5$ & \multicolumn{1}{l|}{IID} & $\alpha=0.5$ \\ \hline
10   & 1 & 50   & 8  & \multicolumn{1}{l|}{1690}   & 3580     & \multicolumn{1}{l|}{550}    & 1200     \\ \hline
10   & 1 & 50   & 32 & \multicolumn{1}{l|}{1750}   & 3640     & \multicolumn{1}{l|}{460}    & 1390     \\ \hline
10   & 1 & 1000 & 8  & \multicolumn{1}{l|}{2810}   & 5820     & \multicolumn{1}{l|}{640}    & 1400     \\ \hline
10   & 1 & 1000 & 32 & \multicolumn{1}{l|}{2710}   & 5490     & \multicolumn{1}{l|}{560}    & 1350     \\ \hline
10   & 4 & 50   & 8  & \multicolumn{1}{l|}{2850}   & 6900     & \multicolumn{1}{l|}{790}    & 1350     \\ \hline
10   & 4 & 50   & 32 & \multicolumn{1}{l|}{3180}   & 7070     & \multicolumn{1}{l|}{750}    & 1830     \\ \hline
10   & 4 & 1000 & 8  & \multicolumn{1}{l|}{4470}   & 9470     & \multicolumn{1}{l|}{1030}   & 1630     \\ \hline
10   & 4 & 1000 & 32 & \multicolumn{1}{l|}{4360}   & 8780     & \multicolumn{1}{l|}{910}    & 1710     \\ \hline
1-19 & 1 & 50   & 8  & \multicolumn{1}{l|}{1710}   & 3730     & \multicolumn{1}{l|}{640}    & 1290     \\ \hline
1-19 & 1 & 50   & 32 & \multicolumn{1}{l|}{2110}   & 3760     & \multicolumn{1}{l|}{500}    & 1520     \\ \hline
1-19 & 1 & 1000 & 8  & \multicolumn{1}{l|}{2770}   & 5950     & \multicolumn{1}{l|}{640}    & 1730     \\ \hline
1-19 & 1 & 1000 & 32 & \multicolumn{1}{l|}{2600}   & 5860     & \multicolumn{1}{l|}{630}    & 1910     \\ \hline
1-19 & 4 & 50   & 8  & \multicolumn{1}{l|}{2930}   & 7780     & \multicolumn{1}{l|}{800}    & 2240     \\ \hline
1-19 & 4 & 50   & 32 & \multicolumn{1}{l|}{2770}   & 5480     & \multicolumn{1}{l|}{800}    & 1900     \\ \hline
1-19 & 4 & 1000 & 8  & \multicolumn{1}{l|}{4470}   & 9060     & \multicolumn{1}{l|}{1130}   & 3100     \\ \hline
1-19 & 4 & 1000 & 32 & \multicolumn{1}{l|}{4100}   & 8640     & \multicolumn{1}{l|}{930}    & 2730     \\ \hline
\end{tabular}
}
\end{table}

\section{Conclusion}
\label{sec:conclusion}

This paper presents the previously unidentified vanishing variance problem encountered in averaging uncorrelated neural network models in gossip learning systems. This problem causes significant convergence delays for model training. By introducing a variance-corrected model averaging algorithm, which involves explicit rescaling of model weights to preserve their original variance, we effectively eliminate the convergence delay associated with this problem. In addition, our averaging algorithm can also be applied to systems that do not suffer from vanishing variance, such as federated learning. Our extensive simulation results show that the fastest node in our algorithm can achieve the same convergence performance as that in federated learning for both IID and non-IID data. Compared to traditional gossip learning, our method is 10x faster than non-optimized gossip learning systems, as well as faster than current state-of-the-art gossip learning systems in all cases. In large-scale networks consisting of 1000 nodes, our method achieves up to 6x faster convergence than current state-of-the-art gossip learning. 

The code and data needed to reproduce the results in this paper are open source and publicly available at: [TODO: FOR DOUBLE BLIND PURPOSE, THIS LINK IS REMOVED] .

% \section*{Declarations}

% \begin{itemize}

% \item Code availability: all code utilized in this paper is open-sourced at [TODO: FOR DOUBLE BLIND PURPOSE, THIS LINK IS REMOVED].

% \item Availability of data and materials: the simulation raw data and the configuration files to reproduce the data are publicly available at [TODO: FOR DOUBLE BLIND PURPOSE, THIS LINK IS REMOVED].

% \item Reproducibility: we note that, while our provide configuration files enable the reproduction of our results, slight variations may occur due to the randomness in model initialization and the random selection of training samples.

% \end{itemize}

%%%%%%%%%%%%%%%%%%%%%%%%%%%%%%%%%%%%%%%%%%%%%%%%%%%%%%%%%%%%%%%%%%%%%%%%

%%% Use this environment to include acknowledgements (optional).
%%% This will be omitted in doubleblind mode.

% \begin{ack}
\section*{Acknowledgment}

\begin{itemize}
    \item This work used the Dutch national e-infrastructure with the support of the SURF Cooperative using grant no. EINF-5527.
    \item This research was performed with the support of the Eureka Xecs project TASTI (grant no.~2022005).
\end{itemize}

% \end{ack}

%%%%%%%%%%%%%%%%%%%%%%%%%%%%%%%%%%%%%%%%%%%%%%%%%%%%%%%%%%%%%%%%%%%%%%%%

%%% Use this command to include your bibliography file.

% \bibliography{ref}

\bibliographystyle{IEEEtran}
\bibliography{ref}

\end{document}